\documentclass[10pt,twocolumn,letterpaper]{article}

\usepackage[pagenumbers]{cvpr} 

\usepackage{graphicx}
\usepackage{amsmath}
\usepackage{amssymb}
\usepackage{booktabs}
\usepackage{csquotes}
\usepackage{mathtools}
\usepackage{booktabs}
 \usepackage{url}

\usepackage[pagebackref,breaklinks,colorlinks]{hyperref}

\usepackage[capitalize]{cleveref}
\crefname{section}{Sec.}{Secs.}
\Crefname{section}{Section}{Sections}
\Crefname{table}{Table}{Tables}
\crefname{table}{Tab.}{Tabs.}

\def\cvprPaperID{59} 
\def\confName{CVPR}
\def\confYear{2022}

\begin{document}

\title{Generate and Edit Your Own Character in a Canonical View} 

\author{Jeong-gi Kwak$^{\dagger}$ \quad Yuanming Li$^{\dagger}$ \quad Dongsik Yoon$^{\dagger}$  \quad David Han$^{\ddagger}$ \quad Hanseok Ko$^{\dagger}$ \\
$^{\dagger}$ Korea University \qquad\qquad $^{\ddagger}$ Drexel Univiersity\\
{}
}
\maketitle

\begin{abstract}
Recently, synthesizing personalized characters from a single user-given portrait has received remarkable attention as a drastic popularization of social media and the metaverse.
The input image is not always in frontal view, thus it is important to acquire or predict canonical view for 3D modeling or other applications. 
Although the progress of generative models enables the stylization of a portrait, obtaining the stylized image in canonical view is still a challenging task.  
There have been several studies on face frontalization but their performance significantly decreases when input is not in the real image domain, e.g., cartoon or painting. Stylizing after frontalization also results in degenerated output. In this paper, we propose a novel and unified framework which generates stylized portraits in canonical view.  With a proposed latent mapper, we analyze and discover frontalization mapping in a latent space of StyleGAN to stylize and frontalize at once. In addition, our model can be trained with unlabelled 2D image sets, without any 3D supervision. 
The effectiveness of our method is demonstrated by experimental results.

\end{abstract}
\vspace{-2mm}
\section{Introduction \& Related work}
\label{sec:intro}

There has been remarkable progress in the fields of image generation and editing via Generative Adversarial Networks (GANs)~\cite{goodfellow2014generative}. The recently proposed GAN models~\cite{karras2018progressive, karras2019style,brock2018large,karras2020analyzing,Karras2021} have advanced to the level of generating photorealistic high-resolution images that humans cannot distinguish, and the concept of GAN has also contributed much to image-to-image (I2I) translation~\cite{Zhu_2017,Choi_2018,huang2018munit}. 


Since the introduction of the method embedding images to StyleGAN's latent space~\cite{abdal2019image2stylegan}, a new paradigm has emerged in image editing. Several techniques of StyleGAN latent manipulation~\cite{shen2020interpreting, harkonen2020ganspace, shen2021closedform} have been combined with GAN inversion~\cite{abdal2019image2stylegan, richardson2021encoding} for real image editing. Because the latent-based editing models~\cite{richardson2021encoding, tov2021designing, abdal2021styleflow} can exploit the rich semantic manifold of StyleGAN, they have shown successful results in altering face attributes (e.g., age, gender, and pose) with high resolution and photorealism.           
\begin{figure}[!t]
\centering \includegraphics[width=\linewidth]{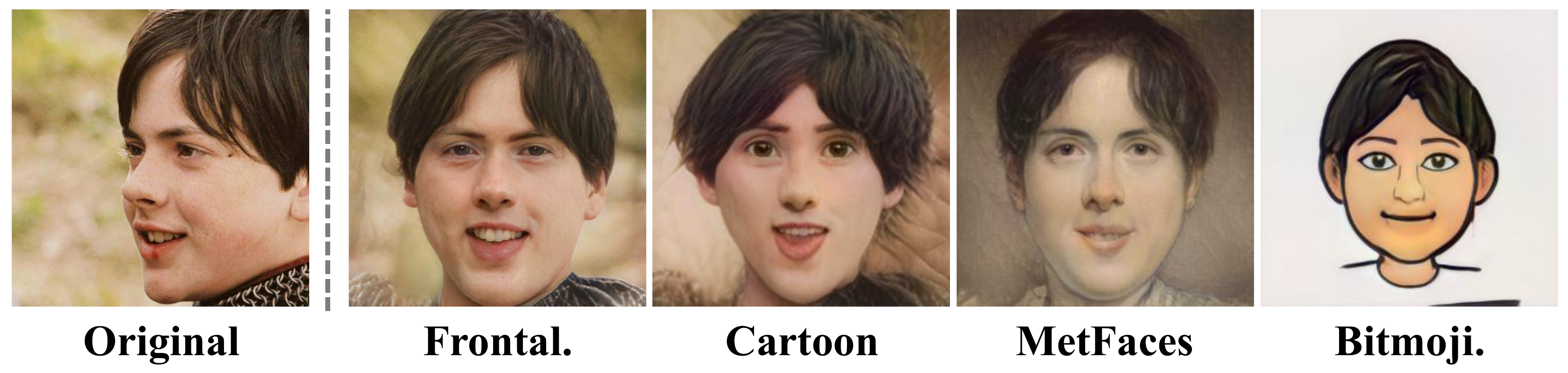}
\caption{Our model can perform stylization and frontalization of an arbitrary image at once without 3D supervision.} 
\label{fig:fig1}
\vspace{-5mm}
\end{figure} 
Beyond semantic editing, the approach can be extended to alter domain via transfer learning of StyleGAN. 
Karras et al.~\cite{Karras2020ada} proposed a method for transfer learning of StyleGAN with a small dataset. Pinkney and Alder~\cite{pinkney2020resolution} suggested an interpolation of two GAN models for controllable image synthesis between different domains, called \enquote{toonify}. Very recently, Song et al.~\cite{song2021agilegan} introduced a novel hierarchical VAE framework for stylizing portraits to various domains. These methods have the potential to be widely used in real-world graphics applications such as movies or cartoons.
\begin{figure*}[!t]
\centering \includegraphics[width=\textwidth]{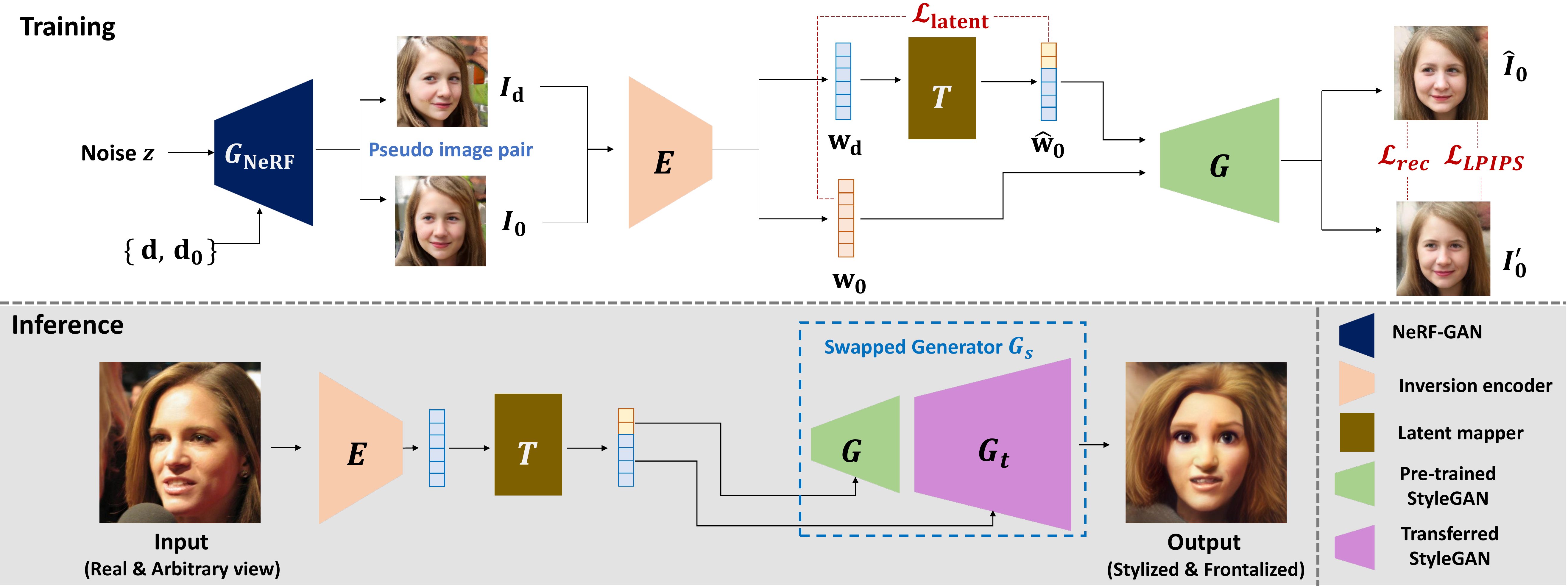}
\caption{The processes about training (top) and inference (bottom) of our model.} 
\label{fig:fig2}
\vspace{-3mm}
\end{figure*} 
However, these methods have limitations in extending to other applications such as game graphics or metaverse. For such tasks, a canonical view of a target should be known, whether it is given or predicted by the model. In reality, a facial image given by a user is not always in frontal view, but rather from an arbitrary angle. However, existing GAN inversion methods are initially aimed to reconstruct the original image thus they cannot synthesize the frontal view of an image taken from a random perspective. Although several  methods~\cite{shen2020interpreting, harkonen2020ganspace, shen2021closedform} have shown the successful  pose control by discovering a direction  related to pose in the latent space of StyleGAN, they haven't found accurate mapping for frontalizing an arbitrary image in an unsupervised manner. InterFaceGAN~\cite{shen2020interpreting} is capable of obtaining canonical pose by using a semantic hyperplane, but it requires supervision (landmark) for binary classification of yaw in order to calculate the hyperplane. 
Several off-the-shelf frontalization methods~\cite{zhou2020rotate, cao2018learning} capable of handling high-resolution images can be considered here to obtain a canonical view of stylized images. However, most face frontalization methods are trained with real facial images, thus their performance would decrease significantly when applied to stylized images because of the domain gap. Application of a similar approach \enquote{frontalize $\rightarrow$ stylize} also presents degenerated results that identities of original images are not well maintained in the output.
\begin{figure}[t]
\centering \includegraphics[width=\linewidth]{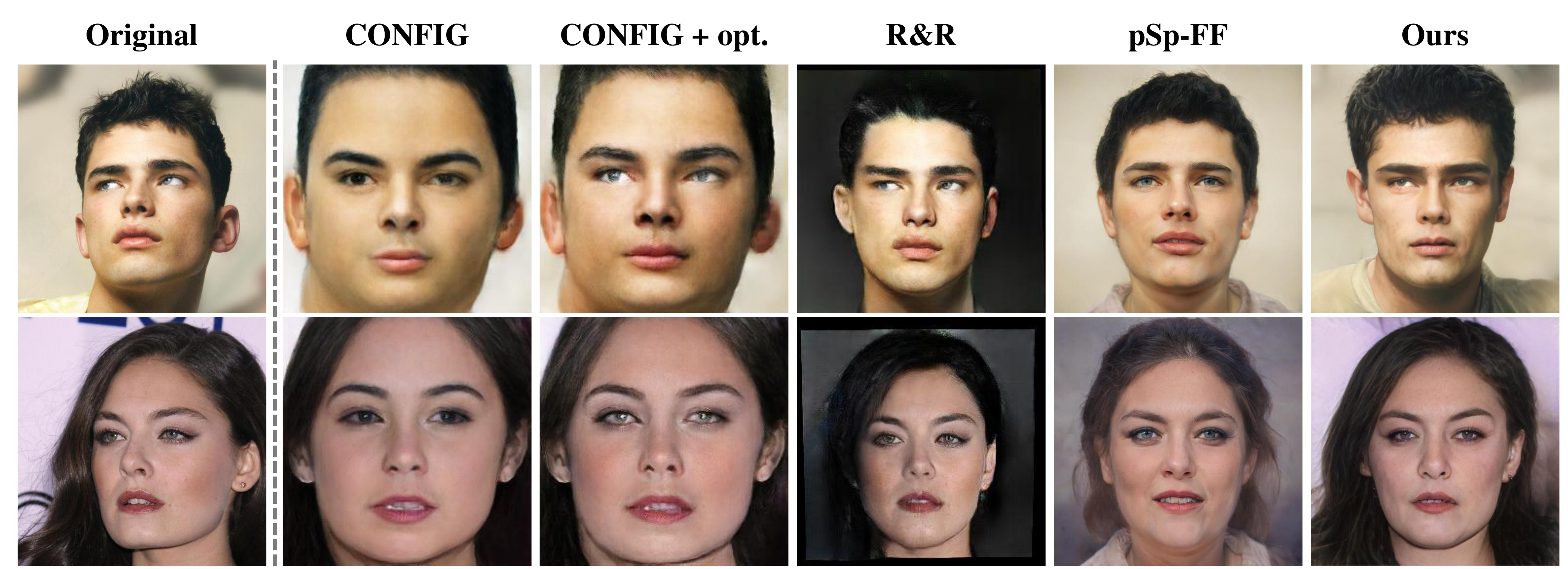}
\caption{Comparison of our method to other   models~\cite{KowalskiECCV2020, zhou2020rotate,richardson2021encoding} which allow face frontalization. pSp-FF is a frontalization version of pSp and note that the results of CONFIG are cropped. } 
\label{fig:fig4}
\vspace{-5mm}
\end{figure} 
To address the above issues, we propose a novel unified framework to obtain a stylized portrait in a canonical view in a single step (Fig.~\ref{fig:fig1}). Our method can discover an accurate latent vector mapping for frontalization without deviating from the StyleGAN manifold. To this end, we propose a novel latent mapper, which is learned to find the elaborate mapping for canonical view with pseudo ground truth pair images generated by NeRF-based GAN~\cite{chan2021pi, mildenhall2020nerf}. Then we utilize the GAN interpolation technique similar to Pinkney and Alder~\cite{pinkney2020resolution}. Here, by designing the manipulated latent code to affect only the former layers of the swapped StyleGAN, we can obtain the stylized image in frontal view without losing visual quality and identity. In addition, the existing methods which handle semantic attribute editing with latent manipulation can be directly applied to ours because the proposed method still follows and preserves the nature of StyleGAN.  
Another notable point is that our model can be learned from unlabelled images, i.e., it requires neither manually annotated labels nor pretrained models such as 3DMM or pose estimator. We demonstrate the effectiveness of our method by qualitative and quantitative comparison with state-of-the-art methods. 
\begin{figure}[t]
\centering \includegraphics[width=\linewidth]{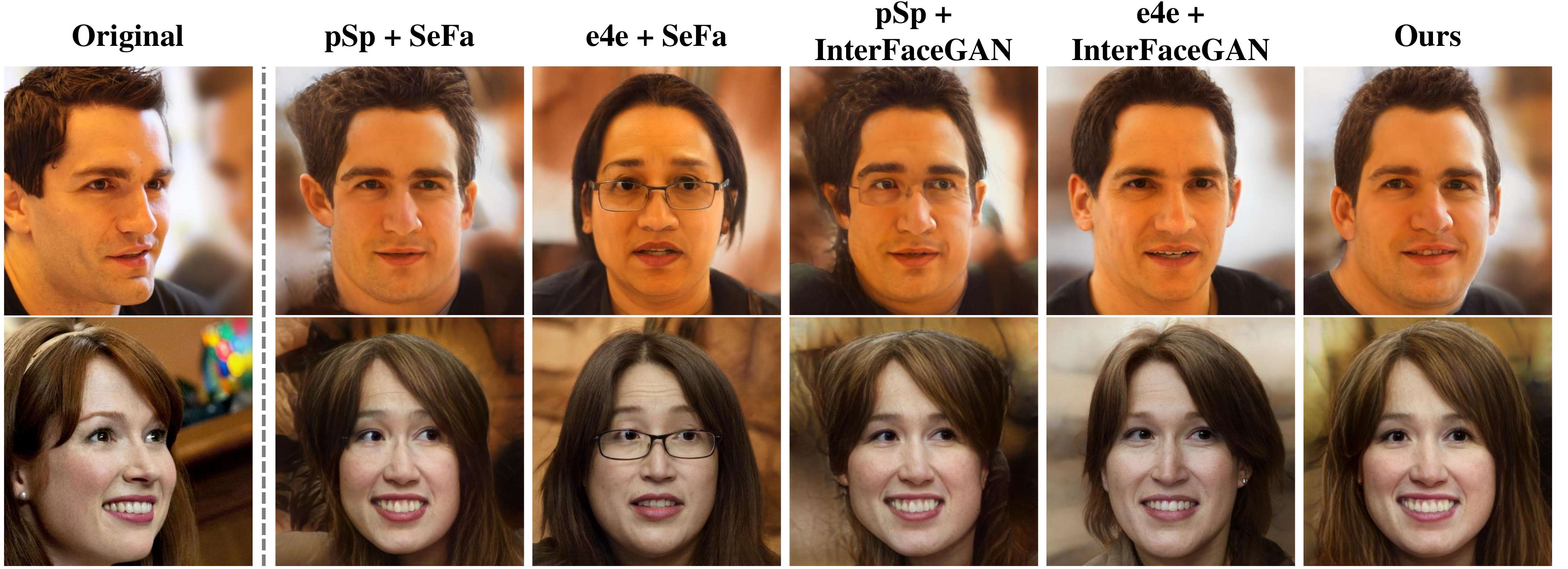}
\caption{Qualitative comparison of our method latent-based manipulation models~\cite{shen2020interpreting,shen2021closedform} with two inversion encoders~\cite{richardson2021encoding,tov2021designing}.} 
\label{fig:fig3}
\vspace{-4mm}
\end{figure} 

\begin{figure*}[!t]
\centering \includegraphics[width=\textwidth]{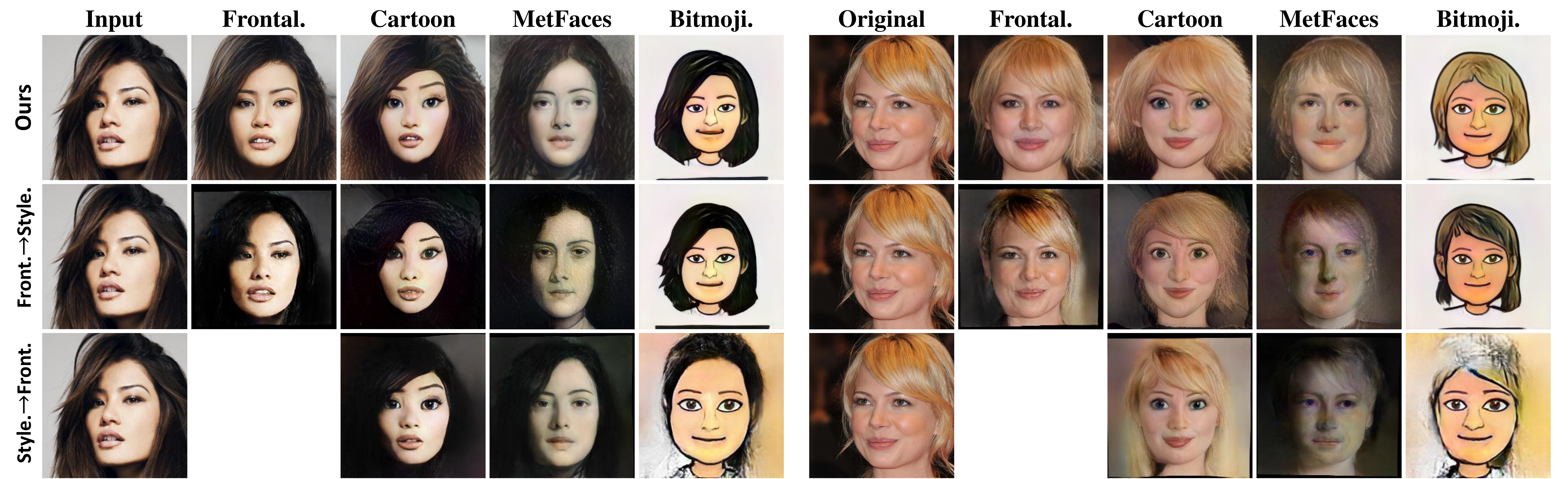}
\caption{Qualitative comparison of our method to cascaded architectures, i.e., \enquote{Frontalization~$\rightarrow$~Stylization} and vice versa. } 
\label{fig:fig5}
\vspace{-1mm}
\end{figure*} 

\section{Method}
\subsection{Frontalization via latent mapping}
Beyond just exploring the direction of related poses, we aim to find an accurate latent mapping to synthesize a frontal view image. We begin with a question: How can we obtain a canonical pose of an arbitrary image without 3D supervision?
To this end, we train a NeRF-based GAN~\cite{chan2021pi} from a scratch that is based on a 3D representation with a neural radiance field~\cite{mildenhall2020nerf} as a generator of  pseudo ground truth pair of random view vs. frontal view images. It learns quite an accurate 3D geometry from 2D images by exploiting a classic volume rendering technique~\cite{kajiya1984ray}. Specifically, it consists of multi-layer perceptron (MLP) which predicts a volume density of the point $\mathbf{\sigma}(\mathbf{x})\in\mathbb{R}$ and the view-dependent RGB color $\mathbf{c}(\mathbf{x}, \mathbf{d})\in\mathbb{R}^3$. Here, $\mathbf{d}$ represents target view direction. After that, the rendered image is obtained by ray-tracing technique~\cite{kajiya1984ray}.
In addition, similar to StyleGAN, the intermediate features are modulated by random noise $\mathbf{z}$. Therefore, the generation procedure of view pair, i.e., random view vs. canonical view, can be described as, 
\begin{equation}
    \{I_{\mathbf{d}},I_{\mathbf{0}}\} = \{G_\text{NeRF}(\mathbf{z}, \mathbf{d}), G_\text{NeRF}(\mathbf{z}, \mathbf{d_0})\}
\end{equation}
where $\mathbf{d}$ and $\mathbf{d_0}$ denote random and frontal  views respectively, and $I_{\mathbf{d}}$ and $I_{\mathbf{d_0}}$ are corresponding images.

However, NeRF-generated images show some artifacts, and resolution is limited because NeRF-GANs require a tremendous amount of memory resource compared to 2D GANs. To overcome the problem, we designed our network to discover a frontalization mapping in the latent space of StyleGAN, not in the image space directly. To this end, we exploit pretrained GAN inversion network~\cite{richardson2021encoding} to embed images into a latent space $\mathcal{W}+$~\cite{abdal2019image2stylegan}. We propose a latent mapper $T$ 
that plays a role in finding a mapping that changes an arbitrary pose to a canonical pose. In training, $I_{\mathbf{d}}$ and $I_{\mathbf{d_0}}$ are embedded to $\mathbf{w_d}$ and $\mathbf{w_0}$ respectively by the inversion encoder $E$, and then $T$ transforms $\mathbf{w_d}$ to $\hat{\mathbf{w}}_{\mathbf{0}}$, i.e., $\hat{\mathbf{w}}_{\mathbf{0}} = T(\mathbf{w_d})$.
$T$ is learned to minimize difference between $\mathbf{w_0}$ and $\hat{\mathbf{w}}_{\mathbf{0}}$ (i.e.,
    $\mathcal{L}_{\text{latent}} = \lVert{\mathbf{w_0}-\hat{\mathbf{w}}_{\mathbf{0}}}\rVert_1$).  

Therefore, the frontalized image of $I_{\mathbf{d}}$ can be obtained by 
$
\hat{I}_{\mathbf{0}}=G(\hat{\mathbf{w}}_{\mathbf{0}}),    
$
where $G$ denotes pretrained StyleGAN generator. 
To guarantee plausible translation result in image space, we also adopt pixel-level $\ell_2$-loss and  LPIPS loss~\cite{zhang2018unreasonable} between two decoded images, i.e.,
\begin{align}
    \mathcal{L}_{\text{rec}} &= \lVert{I'_{\mathbf{0}}-\hat{I}_{\mathbf{0}}}\rVert^2_2, \\
    \mathcal{L}_{\text{LPIPS}} &= \lVert{F(I'_{\mathbf{0}})-F(\hat{I}_{\mathbf{0}})}\rVert^2_2,
\end{align}
where $F(\cdot)$ represents the perceptual feature extractor. Finally, the objective of our method can be formulated as, 
\begin{equation} \label{objective}
     \mathcal{L} = \lambda_{1}\mathcal{L}_{\text{latent}} + \lambda_{2}\mathcal{L}_{\text{rec}} + \lambda_{3}\mathcal{L}_{\text{LPIPS}},
\end{equation}
\noindent where $\lambda_{1}$, $\lambda_{2}$, $\lambda_{3}$ represent the weight of each loss term.
\subsection{Generate stylized image in a canonical view} \label{sec:sec3.2}
As mentioned above, the encoder $E$ encodes an arbitrary image into $\mathcal{W+}$, which is determined by the concatenation of 14 different 512-dimension $\mathbf{w}$ vectors (the case of $256^2$). However, we design that $T$ actually alters only the first four $\mathbf{w}$ vectors (i.e., 4$\times$512) for two reasons. First, several studies have reported that different layers of StyleGAN capture different semantic properties. Among them, \enquote{pose} is related to the early layers. 
Hence, $T$ changes only the parts that affect pose-relevant layers. The second reason is that we leverage GAN interpolation technique~\cite{pinkney2020resolution} to synthesize stylized portraits. As shown in Fig.~\ref{fig:fig2}, instead of conducting transfer learning for all layers, the original StyleGAN (trained with FFHQ) is blended with the transferred StyleGAN (transferred to the target domain). In detail, the swapped model is composed of the original in low-resolution layers ($4^2-8^2$) and the transferred one in all other layers ($16^2-256^2$). 
Here, the latent space related to early layers remains unchanged because they are not swapped.
Thus we can exploit the frontaliztion mapping of $T$ directly for stylized images. As a result, the procedure to transfer an arbitrary image $I$ to  stylized output with canonical view $I_{\text{out}}$ is obtained as  $    I_{\text{out}} = G_{s}(T(E(I)))$,
where $G_{s}$ denotes the swapped generator. 
Because $T(\cdot)$ is still performed in $\mathcal{W}+$ space, the quality is preserved 
and latent manipulation for attribute editing can be applied directly. 
Note that the entire process can be conducted with unlabelled images without any ground-truth or 3D annotations. 
\section{Experiments}
\subsection{Experimental setting}
We use FFHQ dataset~\cite{karras2019style} for training and CelebA-HQ~\cite{karras2018progressive} for test phase, that both datasets consist of high resolution facial images. For NeRF-GAN, we exploit pi-GAN~\cite{chan2021pi} as a multi-view image generator. The resolution of input and output is $256^2$. Also, we exploit pSp encoder~\cite{richardson2021encoding} and StyleGAN2~\cite{karras2020analyzing} as GAN inversion encoder and generator respectively, and they are pre-trained with FFHQ.  To generate stylized portraits, we utilize several datasets in other domains, i.e. cartoon~\cite{pinkney2020resolution}, MetFaces~\cite{Karras2020ada}, and BitmojiFaces~\cite{bitmoji}.
We adopt an adaptive discriminator augmentation (ADA)~\cite{Karras2020ada} for transfer learning.
After that, we interpolate two generators depending on resolution (Fig.~\ref{fig:fig2}). The hyper-parameter of the objective function in Eq.~\ref{objective} are set to $\lambda_{1}$=10 and $\lambda_{2}$=$\lambda_{3}$=1.
All experiments are conducted on a single 24GB RTX 3090 GPU.
\begin{table}[!t] \centering 
\caption{Quantitative comparisons with other approaches.}
\label{tb:tb1}
\begin{tabular}{c c c c }
\toprule
            & Front.~$\rightarrow$~Style. & Style.~$\rightarrow$~Front.  & Ours  \\
 \midrule
FID ($\downarrow$)  &  95.25     &  86.64     & \textbf{65.21}   \\
ID($\uparrow$) & 77.10 & - & \textbf{82.26}\\
Runtime ($\downarrow$) & 1.84  & 4.63 & \textbf{0.26}  \\
\bottomrule
\end{tabular}
\vspace{-3mm}
\end{table}

\subsection{Qualitative comparison}
To demonstrate the effectiveness of the latent mapper, we firstly present the frontalization results in Fig.~\ref{fig:fig4} and Fig.~\ref{fig:fig3}. Fig.~\ref{fig:fig4} shows the comparison with several models~\cite{KowalskiECCV2020,zhou2020rotate,richardson2021encoding} capable of finding a canonical pose. Although CONFIG~\cite{KowalskiECCV2020} (as well as with optimization) and Rotate-and-Render (R\&R)~\cite{zhou2020rotate} use several 3D supervision for training or testing, they suffer from a lack of photorealism and identity-inconsistency.  
pSp-FF, which is a frontalization version of pSp~\cite{richardson2021encoding} can synthesize output without 3D supervision like ours, but loses the identity. 
We also introduce a comparison with the latent-based methods~\cite{shen2020interpreting,shen2021closedform} in Fig.~\ref{fig:fig3}. We exploit two inversion encoders~\cite{richardson2021encoding,tov2021designing} to validate the performance of each model. It can be seen that SeFa~\cite{shen2021closedform} which is an unsupervised method struggles to keep the identity and visual quality. 
InterFaceGAN shows better results than SeFa, and it can generate a frontal pose using a semantic hyperplane, but it needs supervision (landmark) for binary classification to discover the semantic hyperplane.  There may be a method to acquire the hyperplane without supervision by leveraging the concept of flipping image~\cite{wu2020unsupervised}, but it can be applied only to yaw, not other poses such as pitch.
Next, we validate the stylization results of our method.
To our knowledge, there are no approaches that can perform both frontalization and stylization simultaneously. Therefore, for comparison, we exploit the cascaded structures of two independent methods. (R\&R)~\cite{zhou2020rotate} is utilized for face frontalization, and    
we use the same interpolated generator for stylization described in Sec.~\ref{sec:sec3.2}. We divided it in two according to the order of application, i.e. frontalization~$\rightarrow$~stylization and vice versa. As shown in Fig.~\ref{fig:fig5}, both cascaded approaches suffer from a lack of visual quality and losing identity. In contrast, our method delivers successful transfer results without leveraging a heavy off-the-shelf 3D fitting network.
\subsection{Quantitative comparison}
We also present a quantitative comparison of visual quality and computational efficiency. We evaluate FID ~\cite{heusel2017gans} of the transferred image to examine the visual quality of each method. Among the datasets, we use only MetFaces because the others have few images to measure FID. To compare computational efficiency, we evaluate the average runtime to create three different stylized outputs from an input image. Lastly, we report the cosine similarity between original and frontalized images in feature level by ArcFace~\cite{deng2019arcface} to investigate identity preserving (ID). 
The results are listed in Table.~\ref{tb:tb1}, our models achieve a superior FID compared to other methods. Since the frontalization algorithm (R\&R) is computationally expensive due to its 3D fitting and rendering, both \enquote{Front.~$\rightarrow$~Style.} and \enquote{Style.~$\rightarrow$~Front.} approaches are slower than ours. It is remarkable that our approach shows better identity preserving results R\&R which exploits off-the-shelf 3D fitting model~\cite{zhu2017face} (ID in Table.~\ref{tb:tb1}).

\subsection{Editing stylized image in a canonical view}
\begin{figure}[!t]
\centering \includegraphics[width=\linewidth]{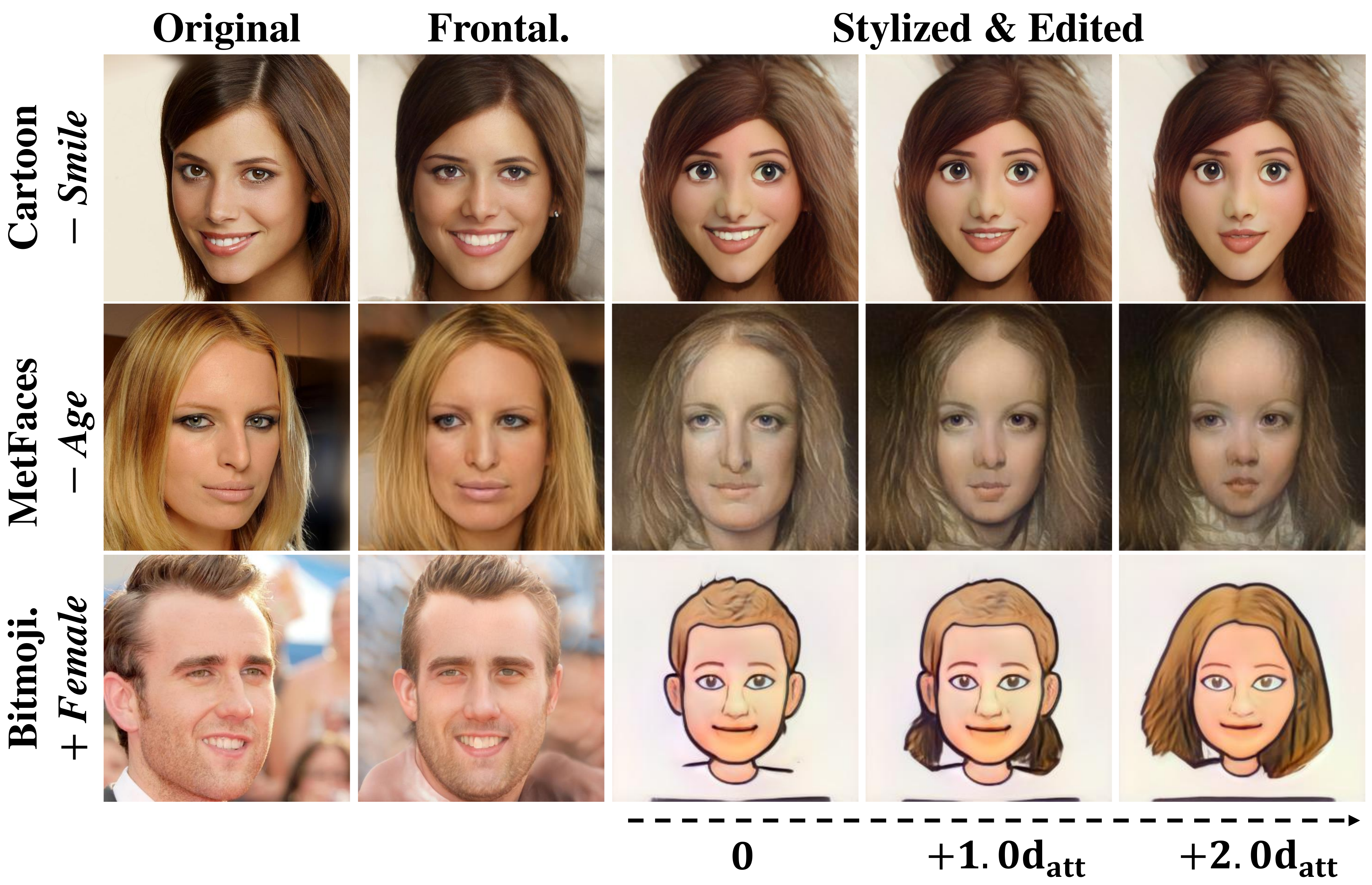}
\caption{Experiment on editing semantic attributes (\textit{smile}, \textit{age}, and \textit{gender})  on stylized portraits in canonical view.} 
\label{fig:fig6}
\vspace{-1mm}
\end{figure}

The manipulation of our model is performed in the latent space of StyleGAN and the edited vector is also placed in the same space. It means that we can apply several existing methods that conduct semantic attribute editing by latent manipulation to our model directly. We exploit InterFaceGAN~\cite{shen2020interpreting} and combine it with ours to edit semantic attributes of stylized images in canonical view. As represented in Fig.~\ref{fig:fig6}, our model successfully alters the attributes (smile, age, and gender) of stylized images. 
\section{Conclusion}

In this paper, we introduced a novel and unified framework to produce stylized portraits in a canonical view. Besides, it can be trained with unlabelled 2D images without any 3D supervision. 
Our method shows superiority in both efficiency and visual quality than applying each method (i.e., frontalization and stylization) successively and it is demonstrated by qualitative and quantitative experiments. We hope our study will have a positive impact on both academia and industry. 

\clearpage

\paragraph{Acknowledgement}
This work was supported by the Brain Korea 21 FOUR Project in 2022.

{\small
\bibliographystyle{ieee_fullname}
\bibliography{egbib}
}

\end{document}